\documentclass{article}

\usepackage{arxiv}

\usepackage[utf8]{inputenc} 
\usepackage[T1]{fontenc}    
\usepackage{hyperref}       
\usepackage{url}            
\usepackage{booktabs}       
\usepackage{amsfonts}       
\usepackage{nicefrac}       
\usepackage{microtype}      
\usepackage{lipsum}

\usepackage{algorithm}
\usepackage{algorithmic}
\usepackage{amsmath}
\usepackage{graphicx}
\usepackage{subcaption}
\usepackage{caption}
\usepackage{xcolor}
\newcommand{\cA}{{\mathcal A}}
\newcommand{\cX}{{\mathcal X}}
\newcommand{\mE}{{\mathbb E}}

\title{Sample-based Distributional Policy Gradient}

\author{
  Rahul Singh \\
  Georgia Tech\\
  \texttt{rasingh@gatech.edu} \\
   \And
 Keuntaek Lee \\
  Georgia Tech\\
  \texttt{keuntaek.lee@gatech.edu} \\
   \And
 Yongxin Chen \\
  Georgia Tech\\
  \texttt{yongchen@gatech.edu} \\
}

\begin{document}
\maketitle

\begin{abstract}
Distributional reinforcement learning (DRL) is a recent reinforcement learning framework whose success has been supported by various empirical studies. It relies on the key idea of replacing the expected return with the return distribution, which captures the intrinsic randomness of the long term rewards. Most of the existing literature on DRL focuses on problems with discrete action space and value based methods. In this work, motivated by applications in robotics with continuous action space control settings, we propose sample-based distributional policy gradient (SDPG) algorithm. It models the return distribution using samples via a reparameterization technique widely used in generative modeling and inference. We compare SDPG with the state-of-art policy gradient method in DRL, distributed distributional deterministic policy gradients (D4PG), which has demonstrated state-of-art performance. We apply SDPG and D4PG to multiple OpenAI Gym environments and observe that our algorithm shows better sample efficiency as well as higher reward for most tasks.
\end{abstract}


\section{Introduction}
\label{sec_introduction}

Reinforcement learning (RL) has shown potential in solving a variety of complex problems in robotics and control~\cite{MniKavDil15,LevFinDar16}. 
RL algorithms can be roughly divided into two categories: value function based and policy gradient methods. Value function based algorithms do not explicitly parameterize the policy, but rather obtain the policy from a learned value function. SARSA~\cite{RumNir94} and Q-learning~\cite{WatDay92} are popular methods for estimation of the value function based on Bellman equation. Recently, deep Q-networks (DQNs)~\cite{MniKavSil15} have been utilized to approximate value function and have demonstrated to achieve human-level performance on computer games. Alternative to value function based approaches, policy gradient methods improve a parameterized policy based on the policy gradient theorem~\cite{SutMcaSin00} and has shown to be more effective in continuous action space control setting. In particular, deep deterministic policy gradient (DDPG)~\cite{LilHunPri16} utilizes neural networks to parameterize the policy and have been successful in solving continuous control tasks.


Instead of modeling the value function as the expected sum of the discounted rewards, recently proposed distributional reinforcement learning (DRL)~\cite{BelDabMun17} framework suggests to work with the full distribution of random returns, known as value or return distribution. Several typical DRL algorithms such as C51~\cite{BelDabMun17}, D4PG~\cite{MarHofBud18}, and QR-DQN~\cite{DabRowBel18} have shown significant performance improvements over non-distributional counterparts in multiple environments including Atari games and DeepMind Control Suite~\cite{Tas18}. 
In DRL, the return distribution is usually represented by discrete categorical form~\cite{BelDabMun17,MarHofBud18,QuManXu18} or quantile function~\cite{DabRowBel18,ZhaMarYao19}. 
Most of existing work within DRL framework are value function based and thus are not suitable for tasks with continuous action space. One of the exceptions is D4PG~\cite{MarHofBud18}, an actor-critic type policy gradient algorithm based on DRL. It has demonstrated much better performance~\cite{MarHofBud18,Tas18} as compared to its non-distributional counterpart (DDPG). However, it still suffers from various drawbacks such as sample inefficiency and extra burden of parameter tuning and projection, which is largely due to the fact that the return distribution in D4PG is modeled by a discrete categorical distribution.

In this paper, we advocate using samples for representing return distribution instead of categorical form or quantiles. Our algorithm which we call sample based distributional policy gradient (SDPG) learns the return distribution by directly generating the return samples via reparameterizing some simple random (e.g. Gaussian) noise samples. SDPG is an actor-critic type policy gradient based algorithm within DRL framework which employs two neural networks: an actor network to parameterize the policy and a critic network to mimic the target return distribution determined via the distributional Bellman equation based on samples. Since the return distribution is usually 1-dimensional, we leverage the quantile Huber loss as a surrogate of the Wasserstein distance for comparing return distributions and thereby learning the critic network.


From a theoretical perspective, SDPG has the following advantages over D4PG:
\begin{itemize}
    \item There is no discretization over the value distributions. The value function network is capable of generating any value distributions, which is in general not categorical as in D4PG.
    \item SDPG does not require the knowledge of the range of the return distribution a prior. In contrast, D4PG requires the domain knowledge in terms of bounds on the return distribution.
    \item No projection is required, instead, Wasserstein distance (quantile Huber loss) gives finer comparison between return distributions.
    \item Once the model is trained, the value distribution can be recovered easily to arbitrary precision by sampling. In contrast, in D4PG, the resolution of the value distribution is fixed once trained.
\end{itemize}
Empirically, we compare the performance of our algorithm with that of D4PG on multiple OpenAI Gym~\cite{BroChePet16} environments for continuous control tasks. We observe that SDPG exhibits better sample efficiency and performs better than or on-par with D4PG in term of rewards in almost all the environments.

\textbf{Related Work:} Most of the algorithms proposed under DRL framework are value function based methods that would run into scalability issue for problems with continuous action space. The C51 algorithm~\cite{BelDabMun17}, a value based algorithm, represented the return distribution using a discrete distribution parameterized by 51 number of uniformly spaced atoms in a specified range. Later, the QR-DQN algorithm~\cite{DabRowBel18} proposed to use discrete set of quantiles to represent the return distribution and demonstrated its effectiveness over C51 algorithm on the Atari 2600 games. QR-DQN was further extended in IQN~\cite{DabOstSilMun18} to learn the full quantile function. D4PG~\cite{MarHofBud18} and Reactor~\cite{GruDabAza17} are existing policy gradient based methods within DRL framework; D4PG dealt with control problems in continuous action space, whereas Reactor was studied in discrete action settings. However, D4PG also utilized the discrete categorical form to represent the return distribution similar to C51 algorithm~\cite{BelDabMun17}, which limits the expressive power of the value distribution network. 
There have been several works on utilizing samples to represent the return distributions. Generative adversarial networks (GANs)~\cite{GodPouMir14}, sample based generative models, have been employed in value function based approaches: GAN-DQN \cite{DoaMazLyl18}, value distribution GAN learning (VDGL)~\cite{FreShiMei19}, and GAN-DDQN~\cite{HuaLiZha19}. GAN-DQN focused on discrete action space and did not show significant improvement over traditional value based methods such as Q-Learning and DQN. Moreover, VDGL utilized GANs to learn multivariate return distributions and thereby learning the value function. Also GAN-DDQN combined GAN and IQN to learn the value function for resource allocation in communication systems. An important point to note is that apart from being value based approaches, the existing GAN based DRL methods employ two networks -- a generator and a discriminator -- for generating return samples by solving a saddle-point problem. In contrast, we utilize quantile Huber loss, as a surrogate of the Wasserstein distance, directly from samples~\cite{DesZhaSch18}, which results in a single objective rather than saddle-point formulation thereby eliminating the need of a discriminator for learning the return distribution.

Rest of the paper is organized as follows. In Section~\ref{sec_background}, we present related background on DRL and sample based modeling of a distribution. The SDPG algorithm is proposed in Section~\ref{sec_algorithm}. Experimental results are presented in Section~\ref{sec_results} followed by the conclusions in Section~\ref{sec_conclusion}.

\section{Background}
\label{sec_background}
\subsection{Distributional RL}
\label{subsec_drl}
We consider a standard RL problem with underlying model $(\cX, \cA, R, P, \gamma)$ where, as usual, $\cX, \cA$ denote the state and action spaces respectively, $R(x,a)$ is the reward of taking action $a$ at state $x$, $P(\cdot \mid x, a)$ is the transition kernel and $0<\gamma<1$ is the discount factor. The reward $R$ can be random in general. This is different to the convention where $R(x,a)$ is the expected reward. The state and action spaces could be either discrete or continuous, though we focus on the more challenging continuous setting. The goal of RL is to find a stationary policy $\pi$ to maximize the long-term accumulated reward 
	\begin{equation}
		\mE \left[\sum_{t=0}^\infty \gamma^t R(x_t, a_t)\right].
	\end{equation}
Since we focus on continuous-state-action tasks, we restrict the policy $\pi$ to be deterministic, that is, $a_t = \pi (x_t)$. 

The Q-function, denoted by $Q^\pi(x,a)$, describes the expected reward of the agent from taking action $a\in \cA$ from state $x \in \cX$, that is,
	\begin{align}
		&Q^\pi(x,a) = \mE \left[\sum_{t=0}^\infty \gamma^t R(x_t, a_t)\right],\\ \nonumber
		&~~x_t \sim P(\cdot\mid x_{t-1}, a_{t-1}), a_t = \pi(x_t), x_0 = x, a_0 = a.
	\end{align}
It satisfies Bellman's equation \cite{Bel66}
	\begin{equation}
		Q^\pi(x,a) = \mE R(x, a) + \gamma \mE Q^\pi (x', \pi(x') \mid x, a). 
	\end{equation}
Here we have adopted the convention that $x'$ denotes the state succeeding $x$, i.e., $x' \sim P(\cdot \mid x, a)$.  

In~\cite{BelDabMun17}, the authors proposed distributional reinforcement learning (DRL), which relies on a random version of Q-function, defined by
	\begin{align}
		&Z^\pi(x,a) = \sum_{t=0}^\infty \gamma^t R(x_t, a_t),\\ \nonumber
		&~~x_t \sim P(\cdot\mid x_{t-1}, a_{t-1}), a_t = \pi(x_t), x_0 = x, a_0 = a.
	\end{align}
Clearly, 
	\begin{equation}\label{eq:QZ}
		Q^\pi(x,a) = \mE Z^\pi(x,a),
	\end{equation}
namely, $Q^\pi$ is the statistical mean of the random variable $Z^\pi$. So in principle, one should be able to recover everything based on $Q^\pi$ using $Z^\pi$. Moreover, the return distribution $Z^\pi$ contains extra information such as variance that may be used to incorporate risk in the RL framework. The $Z$ function satisfies a modified Bellman's equation \cite{BelDabMun17}
	\begin{equation}\label{eq:Bellman}
		Z^\pi(x,a) = R(x,a) + \gamma Z^\pi (x', \pi(x') \mid x, a),
	\end{equation}
where the equation holds in the probability sense. 

Multiple approaches have been proposed to model the return distribution $Z^\pi(x,a)$ including distribution quantiles~\cite{DabRowBel18,ZhaMarYao19} and discrete categorical distribution~\cite{BelDabMun17,MarHofBud18,QuManXu18}. In \cite{BelDabMun17} (D4PG), the return distribution is modeled by a categorical discretization at each $(x,a)$ pair. More specifically, $Z^\pi(x,a)$ is described a probability vector/histogram with fixed bins. The positions of the bins are chosen a prior that need to be tuned according to the environment under consideration.

\subsection{D4PG}
\label{subsec_d4pg}

The DRL was extended to the policy gradient framework in \cite{MarHofBud18}. In policy gradient framework, the policy $\pi$ is modeled by a network $\pi_\theta$ directly. For continuous-state task, a widely used method is deterministic policy gradient (DPG), which relies on the deterministic policy gradient theorem \cite{SilLevHeeRie14}. Let $J(\theta)$ be the average return with control strategy $\pi_\theta$, then
	\begin{equation}
		\nabla J(\theta) = \mE \left[\nabla_\theta \pi_\theta(x) \nabla_a Q^\pi (x,a)\mid_{a=\pi_\theta(x)}\right].
	\end{equation}
This theorem is generalized to the DRL setting \cite{MarHofBud18}, stated as
	\begin{equation}\label{eq:PGE}
		\nabla J(\theta) = \mE \left[\nabla_\theta \pi_\theta(x) \mE[\nabla_a Z^\pi (x,a)]\mid_{a=\pi_\theta(x)}\right].
	\end{equation}
This result follows directly from the fact \eqref{eq:QZ}. The distributed distributional deterministic policy gradients (D4PG) \cite{MarHofBud18} algorithm is based on this extension of policy gradient theorem \eqref{eq:PGE}. It is an actor-critic type algorithm in which the critic learns the return distribution $Z^\pi$ via a neural network. Similar to \cite{BelDabMun17}, the distribution is modeled by a categorical discretization at each $(x,a)$ pair. The actor $\pi_\theta$ is updated via the generalized policy gradient theorem \eqref{eq:PGE} with the expectation being replaced by empirical average.

\subsection{Optimal transport}
\label{OT}

Optimal (mass) transport (OT) is a powerful mathematical tool to study probability distributions \cite{Vil03}. Given two random variable $X, Y$ in the Euclidean space associated with probability distribution $\mu_X, \mu_Y$, the OT problem seeks the solution to 
	\begin{equation}\label{eq:OT}
		\min_{\nu \in \Pi (\mu_X, \mu_Y)} \int c(x,y) \nu(x,y) dxdy
	\end{equation}
with $\Pi (\mu_X, \mu_Y)$ denoting the set of feasible joint distributions of $X$ and $Y$. The unit cost function $c(x,y)$ is often taken to be $||x-y||^p$, $1\le p<\infty$, in which case, \eqref{eq:OT} defines the Wasserstein distance \cite{Vil03} $W_p(\mu_X,\mu_Y)$ between $\mu_X$ and $\mu_Y$. It has an equivalent form 
	\begin{equation}\label{eq:Wp}
		W_p(\mu_X,\mu_Y)^p = \min_{X,Y} \mE ||X-Y||_p^p.
	\end{equation}

The Wasserstein distance is a metric and possesses many nice properties, including the weak continuity \cite{Vil03,Bil71}, which gives reasonable measure of difference between two distributions with disjoint supports. This property is extremely useful in data science as most datasets indeed lie in low-dimensional sub-manifold and therefore any small perturbation can lead to disjoint supports. One representative application relying on this property is the Wasserstein generative adversarial networks \cite{ArjChiBot17}. 

Computing the Wasserstein distance requires solving a linear programming \eqref{eq:OT}. Despite the recent development of algorithms \cite{PeyCut19} in OT algorithms, computation complexity remains a bottleneck of it. One exception is the one-dimensional problem, which has closed-form solution. Let $F_X, F_Y$ be the CDFs of $X, Y$ respectively, then
	\begin{equation}
		W_p(\mu_X, \mu_Y) = \left(\int_0^1 ||F_X^{-1}(u) - F_Y^{-1}(u)||^p du\right)^{1/p}
	\end{equation} 
When only samples generated by $\mu_X, \mu_Y$ are available, then their Wasserstein distance can be approximated as follows. Let $\{x_1, x_2,\ldots, x_n\},\,\{y_1,y_2,\ldots,y_n\}$ be i.i.d. samples corresponding to $\mu_X, \mu_Y$ respectively, and $\{\tilde x_i\}$ ($\{\tilde y_i\}$) be the ascending sorted version of $\{x_i\}$ ($\{y_i\}$), then 
	\begin{equation}\label{eq:sort}
		W_p(\mu_X,\mu_Y)^p \approx \frac{1}{n} \sum_{i=1}^n ||\tilde x_i - \tilde y_i||^p.
	\end{equation}
The approximation error goes to $0$ with rate $1/n$ \cite{DesZhaSch18}. The computational complexity of the sorting operation is $O(n\log n)$. 

However as noted in \cite{DabRowBel18}, Equation~\eqref{eq:sort} does not give unbiased approximation of the Wasserstein distance. In general,
    \[
        {\rm argmin}_{\mu_X}\, \mE[W_p(\mu_X, \hat\mu_Y)]\neq {\rm argmin}_{\mu_X}\, W_p(\mu_X,\mu_Y),
    \]
that is, minimizing the distance to the empirical distribution $\hat\mu_Y$ composed of samples from one distribution is not equivalent to minimizing the distance to that distribution $\mu_Y$ itself \cite{DabRowBel18}. 
To circumvent this difficulty, we borrow tools from quantile regression method and use quantile Huber loss~\cite{Huber1964,DabRowBel18} as a surrogate of the Wasserstein distance. This is given by
\begin{equation}
    \label{eq:huber}
    \frac{1}{n^2} \sum_{i=1}^n \sum_{j=1}^n \rho^{\zeta}_{\hat{\tau}_i} (\tilde x_i - \tilde y_i),
\end{equation}
where
\begin{equation*}
    \rho^{\zeta}_{\hat{\tau}_i} (v) = |\hat{\tau}_i - \delta_{\{v<0\}} | L_\zeta(v),
\end{equation*}
\begin{equation*}
    L_\zeta(v) = \begin{cases} 
    0.5\,v^2 & \text{if~} |v| < \zeta \\
    \zeta(|v| - 0.5~ \zeta) & \text{otherwise,}
    \end{cases}
\end{equation*}
and 
\begin{equation*}
    \hat{\tau}_i = \frac{1}{2} (\tau_i + \tau_{i-1}),~~ i= 1,2,\ldots, n
\end{equation*}
with $\tau_i = \frac{i}{n}$.
\begin{figure*}[h]
\centering
\includegraphics[scale=0.68]{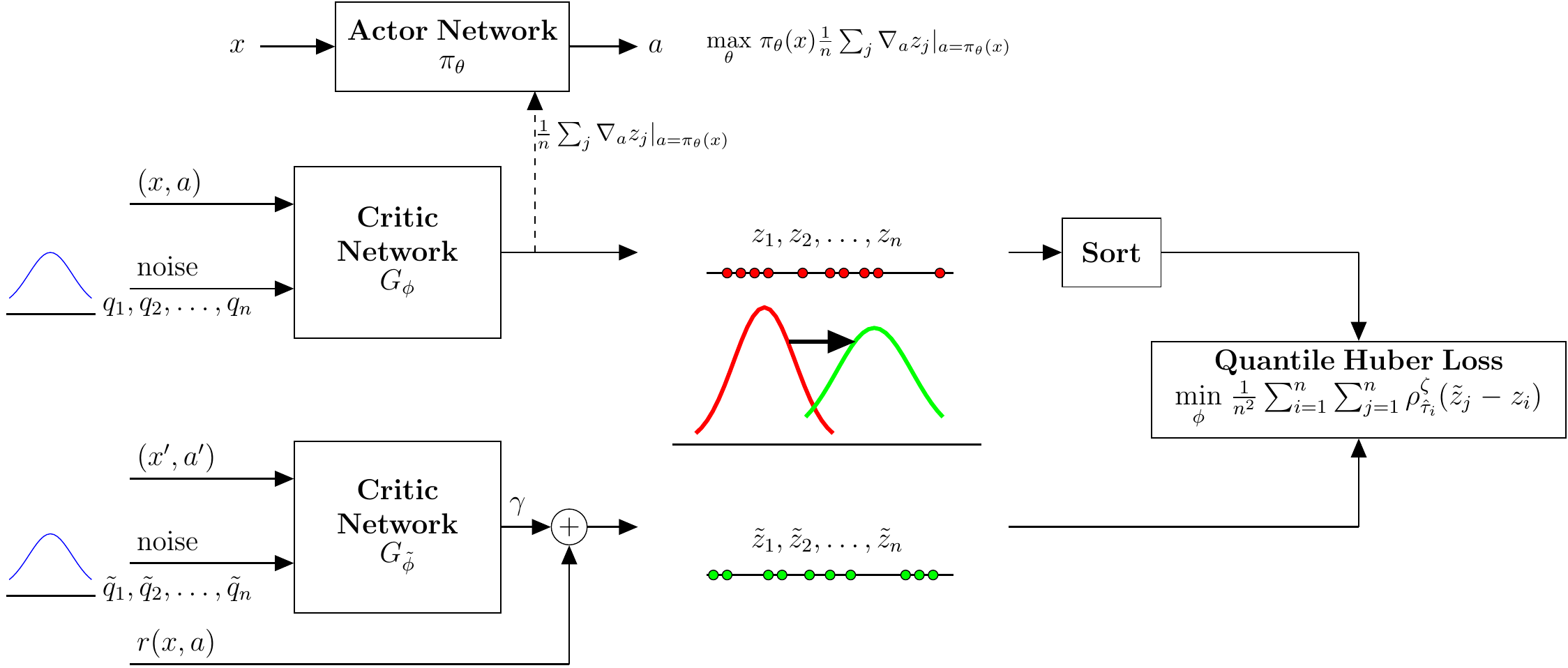}
\caption{Flow diagram of SDPG.}
\label{fig_diagram}
\end{figure*}
\begin{algorithm*}[h]
   \caption{SDPG}
   \label{alg_sdpg}
\begin{algorithmic}
   \STATE {\bfseries Require:} Learning rates $\alpha$ and $\beta$, batch size $M$, sample size $n$, exploration constant $\delta$, 
   \STATE Initialize the the actor network ($\pi$) parameters $\theta$, critic network ($G$) parameters $\phi$ randomly
   \STATE Initialize target networks $(\tilde{\theta}, \tilde{\phi}) \leftarrow (\theta , \phi)$
   \FOR{the number of environment steps}
   
   \STATE Sample $M$ number of transitions $\{(x_t^{i}, a_t^{i}, r_t^{i}, x_{t+1}^{i})\}_{i=1}^M$ from the replay pool
   \STATE Sample noise $\{ q_j^i \}_{j=1}^{n} \sim \mathbb{P}_q $ and $\{ \tilde{q_j}^i \}_{j=1}^{n} \sim \mathbb{P}_q $,~~for~~$i=1,\hdots,M$
   \STATE Apply Bellman update to create samples (of return distribution)
   \begin{align*}
       \tilde{z}_{j}^{i} = r_t^i + \gamma G_{\tilde{\phi}}(\tilde{q_j}^i| (x_{t+1}^i,\pi_{\tilde{\theta}}(x_{t+1}^i))) \quad \mathrm{for}~j=1,2,\hdots, n
   \end{align*}
   \STATE Generate samples $z_j^i = G_{\phi}(q_j^i| (x_{t}^i,a_{t}^i)) \quad \mathrm{for}~j=1,2,\hdots, n $
   \STATE Sort the samples $z^i$ in ascending order
   \STATE Update $G_{\phi}$ by stochastic gradient descent with learning rate $\beta$:
   \begin{align*}
       \frac{1}{M} \sum_{k=1}^{M} \frac{1}{n^2} \sum_{i=1}^{n} \sum_{j=1}^{n} \rho_{\hat{\tau}_i}^{\zeta} (\tilde{z}_{j}^{k} - z_{i}^{k} )
   \end{align*}

   \STATE Update $\pi_{\theta}$ by stochastic gradient ascent with learning rate $\alpha$:
   \begin{align*}
   \frac{1}{M} \sum_{i=1}^{M} \pi_\theta (x_t^i)~ \frac{1}{n} \sum_{j=1}^{n} \left[ \nabla_a z_j^i \right]|_{a = \pi_\theta (x_t^i)}
   \end{align*}
   \STATE Update target networks $(\tilde{\theta}, \tilde{\phi}) \leftarrow (\theta , \phi)$
   \ENDFOR
   \vspace*{0.2cm}
   \STATE {\bfseries Actor}
   \vspace*{-0.2cm}
   
   \hrulefill
   \REPEAT
   \STATE Observe $(x_t,a_t,x_{t+1})$ and draw reward $r_t$
   \STATE Sample action $a_{t+1} = \pi_\theta(x_{t+1}) + \delta \mathcal{N}(0,1)$
   \STATE Store $(x_t,a_t,r_t,x_{t+1},a_{t+1})$ in replay pool
   \UNTIL{learner finishes}

\end{algorithmic}
\end{algorithm*}

\subsection{Reparameterization}
\label{subsec_reparam}
Reparameterization is an effective method to model random variables, especially in case when the goal is to sample from a target distribution instead of modeling them directly using function approximations. Briefly, to model a random variable $X$ with distribution $\mu_X$, reparameterization trick seek a neural network to map a simple random variable $\epsilon$ (e.g. Gaussian) to the target random variable $X$, that is, $X=G(\epsilon)$. The hope is that after training, the random variable $G(\epsilon)$ is closed to $X$ in the probability sense. This is extremely useful for sampling purpose because one only needs to sample from simple distribution $\epsilon$ in order to generate samples of $X$.
The reparameterization trick has been widely used in generative adversarial networks (GANs) \cite{GodPouMir14} where the generator is a map from simple random variable to target data set. It has also been used in variational auto-encoder \cite{KinWel13} to model the encoder. 

\section{Algorithm}
\label{sec_algorithm}

A flow diagram of our SDPG algorithm is shown in Figure~\ref{fig_diagram}. We model the return distribution by samples reparameterized via a simple noise distribution (we use zero-mean Gaussian distribution with unit variance in our implementation). The critic network $G_\phi$, which is a neural network with parameters $\phi$, generates the return samples $z_1,z_2,\hdots, z_n$ for each state and action pair by transforming the noise samples $q_1,q_2,\hdots,q_n$. These generated samples are compared against the samples $\tilde{z}_1,\tilde{z}_2,\hdots,\tilde{z}_n$ generated using the distributional Bellman equation (given by \eqref{eq:Bellman}) employed in the target critic network $G_{\tilde{\phi}}$. The critic network is learned by minimizing the quantile Huber loss as defined in \eqref{eq:huber}, which is a surrogate of the Wasserstein distance between the two 1-dimensional return distributions. Therefore, the loss function for critic network is
\begin{equation}\label{eq:Loss_Critic}
    L_{critic}(\phi) = \mathbb{E} \left[\frac{1}{n^2} \sum_{i=1}^{n} \sum_{j=1}^{n} \rho_{\hat{\tau}_i}^{\zeta} (\tilde{z}_{j} - z_{i} ) \right],
\end{equation}
where $z_1 \geq z_2 \geq \hdots z_n$ are samples after sorting. We emphasize that the sorting is important here to associate each sample with a reasonable $\hat\tau$. This is different to \cite{DabOstSilMun18} where $\hat\tau$ itself is the random seed over $[0\,\,1]$. 

The actor network $\pi_\theta$, parameterized by $\theta$, outputs the action $\pi_\theta(x)$ given a state $x$. The actor network receives feedback from the critic network $G_{\phi}$ in terms of the gradients of the return distribution with respect to the actions determined by the policy. This feedback is used to update the actor network by applying distributional form of the policy gradient theorem given by \eqref{eq:PGE}. Therefore, the gradient of the actor network loss function is
\vspace*{-0.2cm}
\begin{equation}
    \hspace{-0.2cm}\nabla_\theta L_{actor}(\theta)\!\! =\!\! \mathbb{E} \left[ \nabla_\theta \pi_\theta (x)~ \frac{1}{n} \sum_{j=1}^{n} \left[ \nabla_a z_j \right]|_{a = \pi_\theta (x)} \right]
\end{equation}
\vspace*{-0.2cm}

All the steps of our SDPG algorithm are described in Algorithm~\ref{alg_sdpg}. The network parameters of actor and critic networks are updated alternatively in stochastic gradient ascent/descent fashion.

In contrast to the categorical parameterization of the return distribution considered in D4PG, we use samples to represent the return distribution. Since the return distribution is required to be differentiable with respect to the network parameters in order to be learned, we utilize reparameterization trick discussed in Section~\ref{subsec_reparam} to model the return distribution via random noise input. This allows us to learn a continuous distribution via samples as opposed to a discrete-valued categorical distribution in D4PG. Moreover, D4PG requires a projection step in every iteration during training in order to make the target distribution resulting from the distributional Bellman equation coincide with the support of categorical parameterized distribution being learned; SDPG eliminates the need of such a projection step during training. Furthermore, the range of the discretized grid required in D4PG must be tuned according to the reward values for each environment; SDPG does not require such tuning. Another advantage of SDPG is that one can recover the return distribution to arbitrary precision by sampling from the trained critic network. However, the resolution of the return distribution is fixed in D4PG and the critic network has to be trained again from the scratch if one wants to change the resolution. \\

\begin{figure*}[h]	
\vspace*{-0.2cm}
\centering
        \begin{subfigure}[t]{0.12\textwidth}
				\centering
                \includegraphics[width=1.6cm,height=1.6cm]{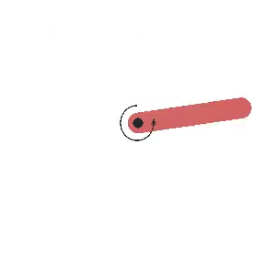}
        \end{subfigure}
        \begin{subfigure}[t]{0.12\textwidth}
				\centering
                \includegraphics[width=1.6cm,height=1.6cm]{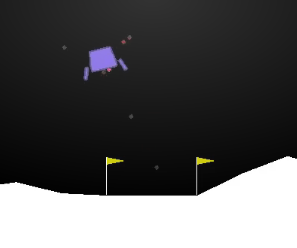}
        \end{subfigure}
        \begin{subfigure}[t]{0.12\textwidth}
				\centering
                \includegraphics[width=1.6cm,height=1.6cm]{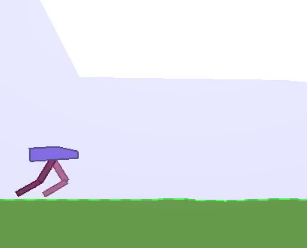}
        \end{subfigure}
        \begin{subfigure}[t]{0.12\textwidth}
				\centering
                \includegraphics[width=1.6cm,height=1.6cm]{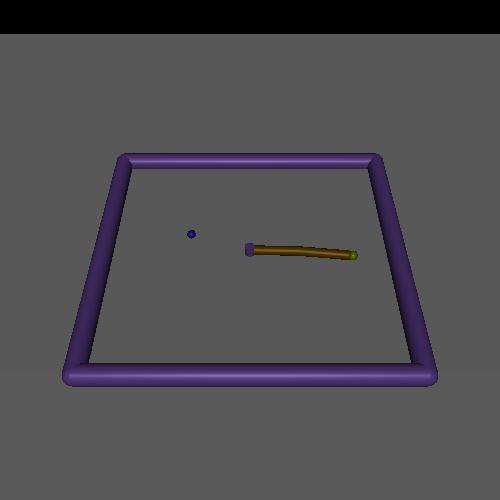}
        \end{subfigure}
        \begin{subfigure}[t]{0.12\textwidth}
				\centering
                \includegraphics[width=1.6cm,height=1.6cm]{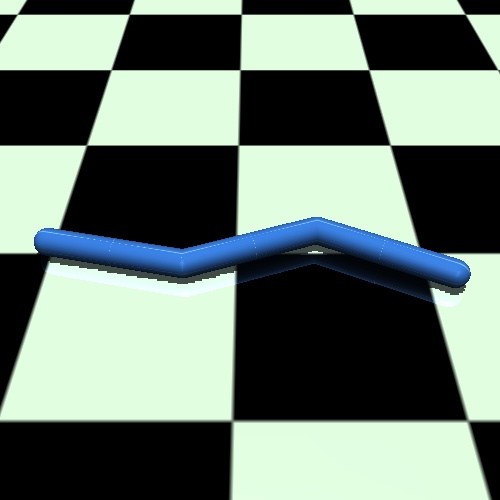}
        \end{subfigure}
        \begin{subfigure}[t]{0.12\textwidth}
				\centering
                \includegraphics[width=1.6cm,height=1.6cm]{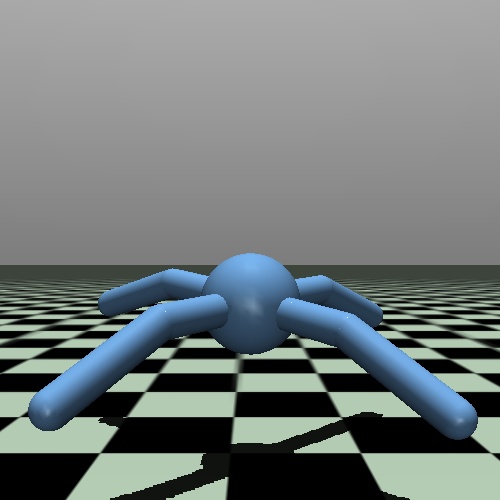}
        \end{subfigure}
        \begin{subfigure}[t]{0.12\textwidth}
				\centering
                \includegraphics[width=1.6cm,height=1.6cm]{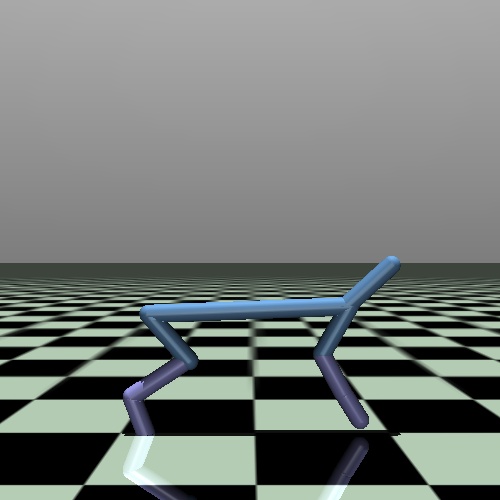}
        \end{subfigure}
        \begin{subfigure}[t]{0.12\textwidth}
				\centering
                \includegraphics[width=1.6cm,height=1.6cm]{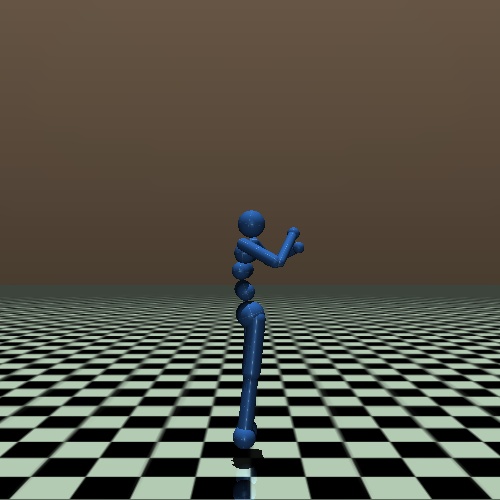}
        \end{subfigure}
        \caption{Illustration of each OpenAI Gym environment. From left to right: Pendulum-v0, LunarLanderContinuous-v2, BipedalWalker-v2, Reacher-v2, Swimmer-v2, Ant-v2, HalfCheetah-v2, Humanoid-v2.}
        \label{fig_sample_env}
\end{figure*}

\section{Experimental Results}
\label{sec_results}

We compare the performance of proposed SDPG with D4PG algorithm on a range of challenging continuous control tasks from the OpenAI Gym environments. Figure~\ref{fig_sample_env} shows example screenshots of samples from different domains considered in our experiments. Note that in the original D4PG paper, the environments considered were from DeepMind Control Suite~\cite{Tas18} where the rewards are bounded between 0 to 1 for all the domains. In \cite{BelDabMun17} and \cite{Tas18}, it was demonstrated that D4PG outperforms DDPG~\cite{LilHunPri16} in almost all the environments and therefore, we compare our algorithm with the only existing policy gradient method in DRL -- D4PG. All the experiments are performed using TensorFlow with one NVIDIA TITAN Xp GPU. 

For both actor and critic networks, we use a two layer feedforward neural network with hidden layer sizes of 400 and 300, respectively, and rectified linear units (ReLU) between each hidden layer. We also used batch normalization on all the layers of both networks. Moreover, the output of the actor network is passed through a hyperbolic tangent (Tanh) activation unit. 

\begin{figure}[h]	
\centering
\hspace*{-1cm}
    \includegraphics[scale=0.21]{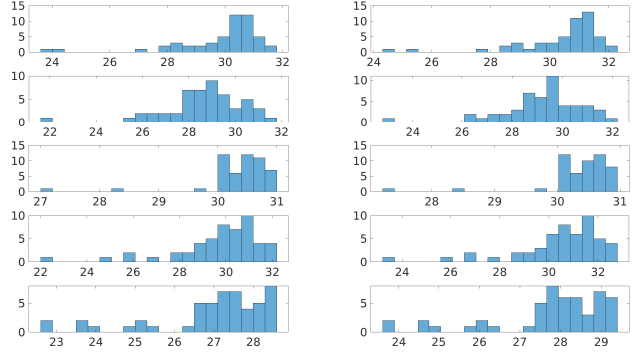}
\caption{Histogram of the 51 samples from random state-action pairs after 1 million iterations on BipedalWalker-v2 domain: left column shows the histograms of samples generated by the critic network and right column shows the corresponding histograms after applying the distributional Bellman equation. Note that the learned critic network can be used to represent the return distribution at arbitrary resolution by generating as many samples as required.}
\label{fig_histogram_walker_51}
\end{figure}
\begin{figure}[h]	
\centering
\hspace*{-1cm}
    \includegraphics[scale=0.21]{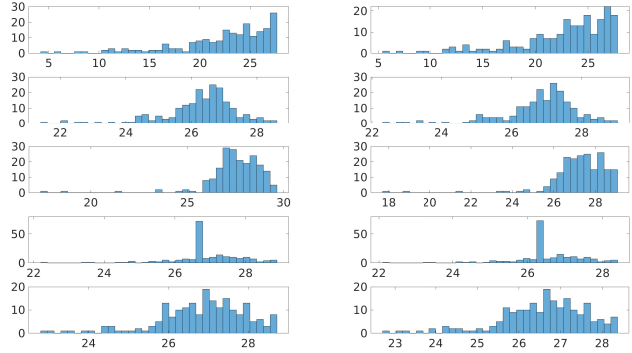}
\caption{Histogram of the 200 samples. The critic network was trained with 51 number of samples and the same network was used to generate these 200 samples.}
\label{fig_histogram_walker_200}
\end{figure}

\begin{figure}[h]	
\centering
    \includegraphics[scale=0.3]{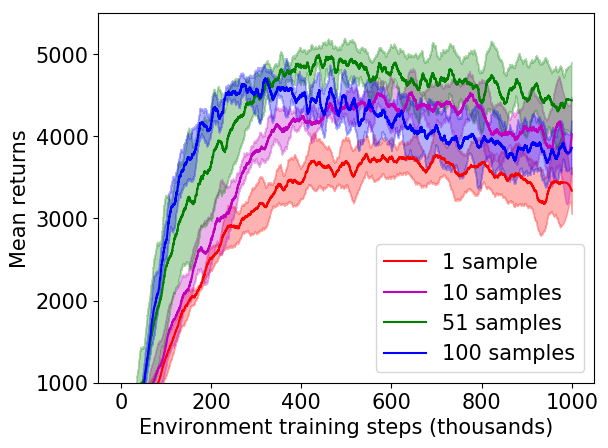}
\caption{Effect of number of samples on Ant-v2.}
\label{fig_ant_atoms}
\end{figure}

In all experiments we use learning rates of $\alpha = \beta = 1 \times 10^{-4}$, batch size $M=256$, exploration constant $\delta = 0.3$, and $\zeta = 1$. We use a replay table of size $R = 1 \times 10^6$ for all the domains except $R=0.2 \times 10^6$ for Pendulum and LunarLanderContinuous. Across all the tasks, for D4PG we use 51 atoms to represent the categorical distribution and similarly, for SDPG we use $n =51$ number of samples to represent return distributions. Moreover, we run each task for a maximum of 1000 steps per episode. Note that SDPG is a centralized algorithm at this moment for a single agent. Thus, the distributed algorithm in D4PG is deactivated for fair comparison. One can easily establish a distributed version of SDPG. Since SDPG requires sorting operation during training, SDPG takes a little more time per episode (almost $\times 1.3$) than D4PG.

\begin{figure*}[h]	
\centering
        \begin{subfigure}[t]{0.24\textwidth}
				\centering
                \includegraphics[scale=0.28]{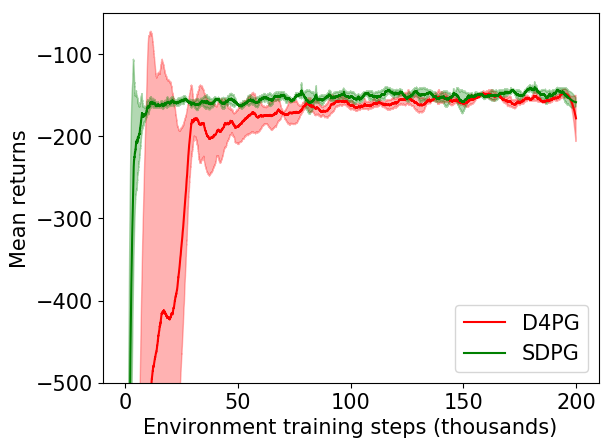}
                \caption{Pendulum-v0.}
        \end{subfigure}\hspace*{0.2cm}
        \begin{subfigure}[t]{0.24\textwidth}
				\centering
                \includegraphics[scale=0.28]{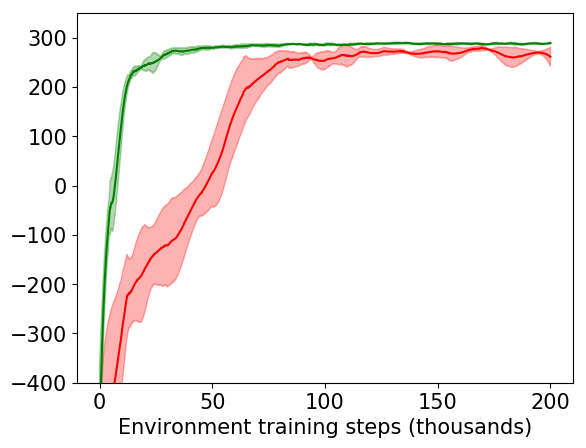}
                \caption{LunarLander-v2.}
        \end{subfigure}\hspace*{0.2cm}
        \begin{subfigure}[t]{0.24\textwidth}
				\centering
                \includegraphics[scale=0.28]{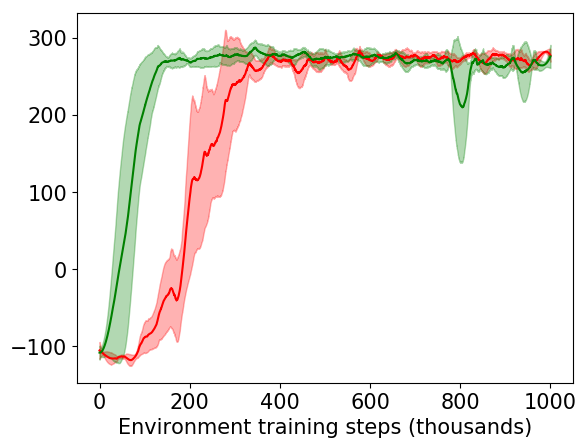}
                \caption{BipedalWalker-v2.}
        \end{subfigure}\hspace*{0.2cm}
        \begin{subfigure}[t]{0.24\textwidth}
				\centering
                \includegraphics[scale=0.28]{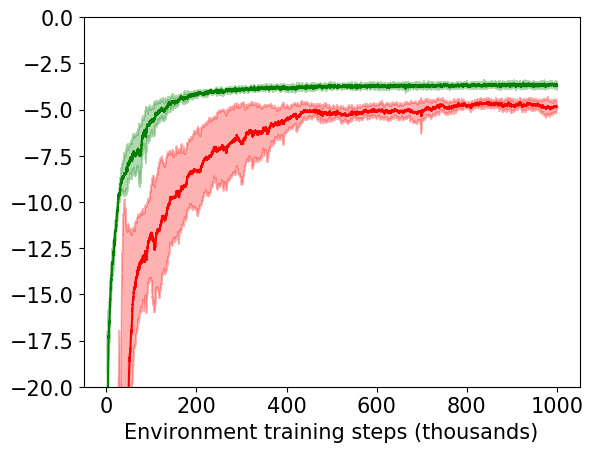}
                \caption{Reacher-v2.}
        \end{subfigure}\\
        \begin{subfigure}[t]{0.24\textwidth}
				\centering
                \includegraphics[scale=0.28]{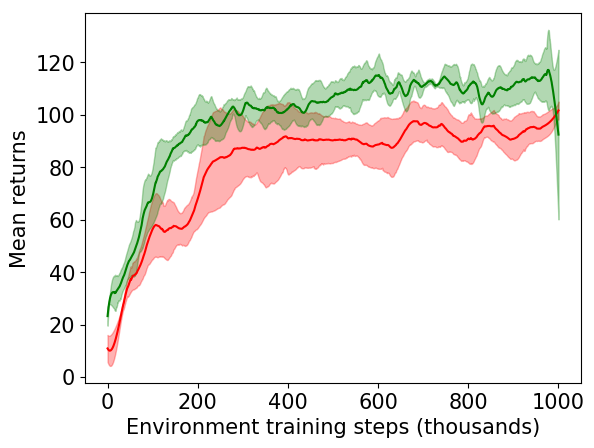}
                \caption{Swimmer-v2.}
        \end{subfigure}\hspace*{0.2cm}
        \begin{subfigure}[t]{0.24\textwidth}
				\centering
                \includegraphics[scale=0.28]{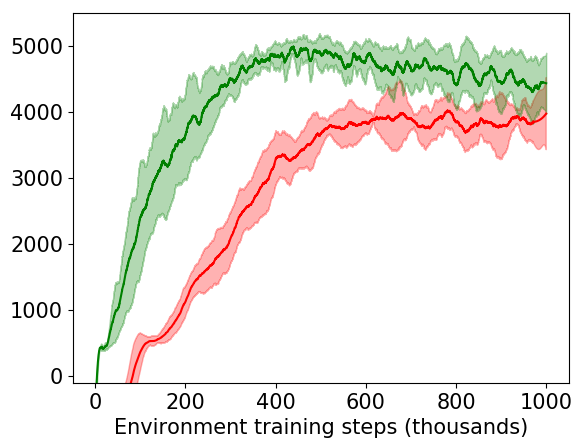}
                \caption{Ant-v2.}
        \end{subfigure}\hspace*{0.2cm}
        \begin{subfigure}[t]{0.24\textwidth}
				\centering
                \includegraphics[scale=0.28]{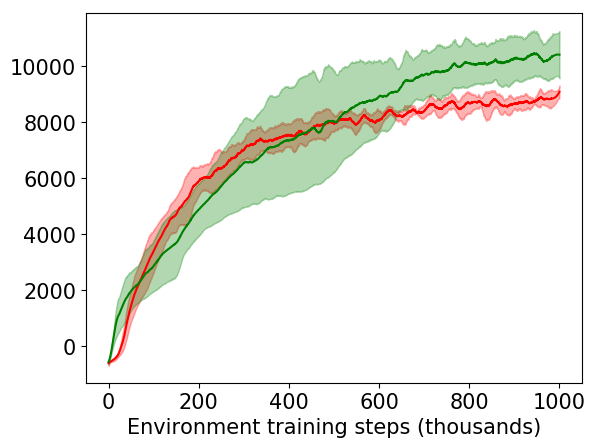}
                \caption{HalfCheetah-v2.}
        \end{subfigure}\hspace*{0.2cm}
        \begin{subfigure}[t]{0.24\textwidth}
				\centering
                \includegraphics[scale=0.28]{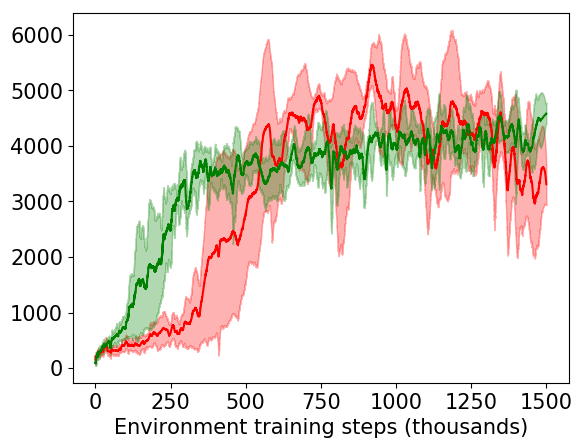}
                \caption{Humanoid-v2.}
        \end{subfigure}
        \caption{Mean returns of SDPG (green color) and D4PG (red color) agents over environments steps for different OpenAI Gym environments. The shaded region represents standard deviation of the average returns over 5 random seeds. The curves are smoothed uniformly for visual clarity.}
        \label{fig_compare}
\end{figure*}

\begin{table*}[h]
\vspace*{-0.4cm}
  \caption{Comparison of average maximal returns $\pm$ one standard deviation over 5 different trials. The evaluations are done every 5000 environment steps in each trial over 100 episodes.}
  \label{table_1}
  \centering
  \begin{tabular}{lllll}
    \toprule
    \multicolumn{4}{r}{Reward}                   \\
    \cmidrule(r){4-5}
    Domain & Simulator & Train Steps    & D4PG     & SDPG \\
    \midrule
    Pendulum & Classic control & $0.2\times 10^6$  & $-144.63 \pm 6.72$  & $\mathbf{-142.46 \pm 5.21}$     \\
    LunarLander  & Box2D   & $0.2\times 10^6$ &  $283.22 \pm 1.63$ & $\mathbf{291.71 \pm  0.90}$      \\
    BipedalWalker &  Box2D  & $1.0\times 10^6$    & $ 289.66 \pm 3.26$  & $\mathbf{301.91 \pm 4.38}$   \\
    Reacher & MuJoCo    & $1.0\times 10^6$    & $-4.40 \pm 0.28$  & $\mathbf{-3.62 \pm 0.20}$   \\
    Swimmer  & MuJoCo    & $1.0\times 10^6$    & $115.52 \pm 4.94$  & $\mathbf{128.38 \pm 6.20}$   \\
    Ant  & MuJoCo   & $1.0\times 10^6$    & $5058.84 \pm 93.57$  & $\mathbf{5762.83 \pm 194.67}$   \\
    HalfCheetah  & MuJoCo   & $1.0\times 10^6$    & $9565.06 \pm 209.77$  & $\mathbf{10607 \pm 845.04}$   \\
    Humanoid  & MuJoCo   & $1.5\times 10^6$    & $\mathbf{6064.14 \pm 192.84}$  & $5093.62 \pm 402.25$   \\
   
    \bottomrule
  \end{tabular}
\end{table*}
\subsection{Training and Evaluation}
\label{subsec_training}
First we demonstrate the ability of the critic network to learn the return distribution utilizing the variant of Huber loss as in Equation~\eqref{eq:huber} based on the distributional Bellman equation. Figure~\ref{fig_histogram_walker_51} shows the histograms of the samples generated by the learned critic network and the corresponding histograms of samples generated based on the distributional Bellman equation on BipedalWalker-v2 domain. Clearly, the histograms match almost perfectly which demonstrates that the critic network in SDPG successfully learns the target return distribution determined via the distributional Bellman equation. Note that this learned critic network can be used to generate as many samples as required (see Figure \ref{fig_histogram_walker_200}) to approximate the return distribution at arbitrary resolution.\\

Next, we study the effect of varying number of samples representing the return distribution while training. Figure~\ref{fig_ant_atoms} depicts the training curves with different samples on Ant-v2 domain. For a fixed number of samples, the algorithm is trained for five different seeds: the solid lines represent the mean returns over five trials and the shaded region represent the corresponding standard deviation. 
Initially, increasing the number of samples improves the performance in terms of efficiency as well as returns. However, when using 100 samples for training although an improvement in efficiency is observed, the returns have gone down significantly.

For comparison, we train five different instances of each algorithm with different random seeds, with each performing one evaluation rollout every 5000 environment steps. Figure~\ref{fig_compare} shows the comparison of mean returns on different environments.
It is evident form the figure that SDPG exhibits significantly better sample efficiency than D4PG on almost all the environments. Moreover in terms of average returns, SDPG performs better than D4PG on all the domains except Humanoid-v2. This maybe due to insufficient training steps. The performance of SDPG for Humanoid-v2 keeps increasing during the entire training process and this trend is expected to continue.

\begin{table}[h]
\vspace*{-0.4cm}
  \caption{Comparison of sample efficiency in terms of number of episodes needed to reach a certain threshold. The episodes reported here are the smallest $m$ for which the mean episode reward over $m^{th}$ and $(m+10)^{th}$ episodes crosses a certain return threshold. The number of episodes are averaged over 5 different trials. The thresholds are chosen according to~\cite{GuLilGha17}.}
  \label{table_2}
  \centering
  \begin{tabular}{llll}
    \toprule
    \multicolumn{4}{r}{Episodes}                   \\
    \cmidrule(r){3-4}
    Environment & Threshold    & D4PG     & SDPG \\
    \midrule
    Pendulum & -150  & $1605$  & $\mathbf{481}$     \\
    LunarLander    & 200 &  $2994$ & $\mathbf{1039}$      \\
    BipedalWalker   & 250   & $3039$  & $\mathbf{1479}$   \\
    Reacher     & -7    & $36623$  & $\mathbf{11859}$   \\
    Swimmer     & 90    & $\mathbf{3492}$  & $3867$   \\
    Ant     & 3500    & $5567$  & $\mathbf{4437}$   \\
    HalfCheetah     & 4700    & $\mathbf{3217}$  & $7692$   \\
    Humanoid     & 2500    & $57142$  & $\mathbf{52749}$   \\
    \bottomrule
  \end{tabular}
\end{table}


We evaluate the performance of our algorithm, SDPG, based on two criteria: average returns and sample efficiency. Table~\ref{table_1} lists the maximal mean returns (the average of the maximal returns over different trials) along with the standard deviation. The average returns are evaluated every 5000 training steps over 100 episodes. We observe that the returns for SDPG are significantly better than D4PG for all the environments except Humanoid. 
To compare the sample efficiency of D4PG and SDPG, we list the number of episodes needed to reach certain return threshold in Table~\ref{table_2}. The episode numbers reported in the table are averaged over 5 different trials and for each trial the episode number is the number of episodes required before the reward crosses a certain threshold. It is evident that SDPG requires significantly smaller number of episodes than D4PG on many environments.

\section{Conclusion}
\label{sec_conclusion}
In this paper, driven by applications in continuous action space, we proposed sample-based distributional policy gradient (SDPG) algorithm for learning the policy within DRL framework. This algorithm is a combination of an actor-critic type of policy gradient method and DRL. Departing from the existing state-of-art distributional policy gradient algorithm D4PG, the sampled-based reparameterization technique used in SDPG enables us to learn the return distribution to arbitrary resolution. We compared the performance of SDPG with D4PG on multiple OpenAI Gym environments. Our algorithm showed better sample efficiency than D4PG in most environments and performed better than D4PG in terms of average returns.

\bibliographystyle{unsrt}  
\bibliography{sdpg}  


\end{document}